\documentclass[letterpaper]{article} 
\usepackage{aaai2026}  
\usepackage{times}  
\usepackage{helvet}  
\usepackage{courier}  
\usepackage[hyphens]{url}  
\usepackage{graphicx} 
\usepackage{amssymb} 
\usepackage{natbib}  
\usepackage{caption} 
\usepackage{algorithm}
\usepackage{algorithmic}
\usepackage{textcomp} 

\usepackage{amsmath}
\usepackage{booktabs}
\usepackage{multirow}
\usepackage{pifont}
\usepackage{makecell}
\usepackage{textcomp}
\usepackage{placeins}
\usepackage[table]{xcolor}

\usepackage{newfloat}
\usepackage{listings}
\DeclareCaptionStyle{ruled}{labelfont=normalfont,labelsep=colon,strut=off} 
\floatstyle{ruled}
\newfloat{listing}{tb}{lst}{}
\floatname{listing}{Listing}
\title{PipeDiT: Accelerating Diffusion Transformers in Video Generation with Task Pipelining and Model Decoupling}
\author{
    Sijie Wang,
    Qiang Wang,
    Shaohuai Shi\thanks{Corresponding author.}
}

\affiliations{
    School of Computer Science and Technology, Harbin Institute of Technology, Shenzhen\\

    25b951105@stu.hit.edu.cn, \{qiang.wang, shaohuais\}@hit.edu.cn
}

\begin{document}

\maketitle

\begin{abstract}
Video generation has been advancing rapidly, and diffusion transformer (DiT) based models have demonstrated remarkable capabilities. However, their practical deployment is often hindered by slow inference speeds and high memory consumption. In this paper, we propose a novel pipelining framework named PipeDiT to accelerate video generation, which is equipped with three main innovations. First, we design a pipelining algorithm (PipeSP) for sequence parallelism (SP) to enable the computation of latent generation and communication among multiple GPUs to be pipelined, thus reducing inference latency. Second, we propose DeDiVAE to decouple the diffusion module and the variational autoencoder (VAE) module into two GPU groups, the executions of which can also be pipelined to reduce memory consumption and inference latency. Third, to better utilize the GPU resources in the VAE group, we propose an attention co-processing (Aco) method to further reduce the overall video generation latency. We integrate our PipeDiT into both OpenSoraPlan and HunyuanVideo, two state-of-the-art open-source video generation frameworks, and conduct extensive experiments on two 8-GPU systems. Experimental results show that, under many common resolution and timestep configurations, our PipeDiT achieves $1.06\times$ to $4.02\times$ speedups over OpenSoraPlan and HunyuanVideo.
\end{abstract}


\section{Introduction}

\begin{figure}[!t]
\centering
\includegraphics[width=\columnwidth]{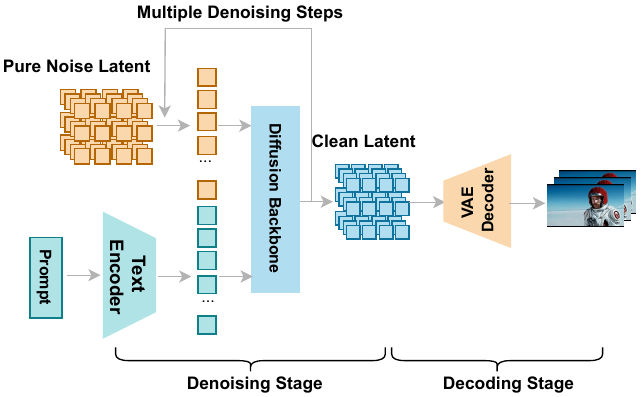} 
\caption{Text-to-video generation starts with encoding the input text and a pure noise latent into a semantic representation, which guides a diffusion model to iteratively refine a latent. The refined latent is then upsampled by a VAE decoder to generate the final video.}
\label{fig:dit-process}
\end{figure}

Video generation models~\cite{li2024survey,cho2024sora,sun2024sora,blattmann2023stable}
 have advanced rapidly over the past two years. Such models take textual or visual inputs and synthesize a continuous video as output. 
 Diffusion Transformers (DiT)~\cite{fan2025vchitect,ma2024latte,yang2024cogvideox} have emerged as the primary framework for video generation due to the high quality of the videos they produce.
 The key idea of DiT is to adopt a progressive denoising process~\cite{ho2020denoising,sohl2015deep,song2019generative}: starting from pure noise, the model iteratively refines the signal through a reverse diffusion process until a high-quality video is generated as shown in Fig.~\ref{fig:dit-process}. 
 However, the inherently sequential nature of its reverse diffusion process severely restricts parallelism during inference~\cite{li2023snapfusion,shih2023parallel,chen2024pixart}. 
 
 Current inference optimizations for DiT target both image and video generation~\cite{liu2024linfusion,zhang2024partially,luo2025accelerating}. For image generation, DistriFusion\cite{li2024distrifusion} accelerates inference by splitting the input into multiple patches and distributing them across different GPUs. It reuses intermediate feature maps from the previous timestep to provide context for the current step, and hides communication overhead via asynchronous communication in the computation pipeline. PipeFusion~\cite{fang2024pipefusion} also divides images into patches and distributes network layers across multiple GPUs to address memory limitations during generation. For video generation, methods~\cite{chen2024delta,selvaraju2024fora} like Teacache~\cite{liu2025timestep} analyze the correlation between features across adjacent timesteps and reuse outputs from the previous step to reduce the number of timesteps, thus improving inference efficiency~\cite{ma2024deepcache,zhao2024real}. However, these approaches may theoretically introduce degradation in generation quality. Consequently, the majority of current video generation models utilize system-level optimizations such as sequence parallelism (SP) \cite{li2023lightseq,sun2024linear,zhao2024dsp} to expedite the generation process while preserving the quality of the generated videos.

Currently, two main SP paradigms have been proposed. The first is DeepSpeed-Ulysses (termed as Ulysses afterward)~\cite{jacobs2023deepspeed}. By splitting the attention heads and forming complete Query ($\mathbf{Q}$), Key ($\mathbf{K}$) and Value ($\mathbf{V}$) sequences, Ulysses parallelizes attention calculation across multiple GPUs. Its primary advantage lies in its communication pattern, which involves three All-to-All operations before attention computation and one afterward, resulting in relatively low communication overhead. 
\textit{However, its scalability is limited by the number of attention heads, and existing implementations do not overlap computation and communication, leaving GPU resources underutilized.}
The second is Ring-Attention~\cite{li2021sequence}, which performs local attention computations on a partial sequence, and then gathers the \textbf{K} and \textbf{V} tensors across all devices using Peer-to-Peer (P2P) communication to complete global attention. This method supports much higher degrees of parallelism, \textit{but the increased communication overhead can negate its benefits}, making Ulysses generally preferable when parallelism is not inherently restricted. Some prior works have combined both methods into Unified Ulysses-Ring SP (USP)~\cite{fang2024usp,fang2024xdit}, mitigating the limited parallelism of Ulysses at the cost of additional communication overhead. 

While current system-level optimizations aim to speed up DiT-based video generation inference reserving the video quality, they still face two key issues: 1) communication between GPUs during the denoising phase often hampers inference efficiency, and 2) using a decoding VAE can quickly result in out-of-memory (OOM) errors, rendering decoding inefficient.
In this paper, we introduce PipeDiT, a system-level optimized inference framework, which employs three innovative methods to reduce video generation latency while maintaining video output quality.
First, we design a pipeline algorithm, named PipeSP, that overlaps communication and computation within Ulysses to hide some communication overhead and improve GPU utilization. 
Second, to address the GPU memory explosion in the decoding stage caused by colocation, we propose DeDiVAE to decouple the diffusion module and VAE decoder and onto two GPU groups. DeDiVAE greatly reduces peak GPU memory usage while allowing pipelined execution of decoding and denoising computations. 
Third, to address the suboptimal utilization of some GPUs caused by DeDiVAE, we further propose an attention co-processing (Aco) module which breaks down DiT into its linear-layer and attention-computation components. This fine-grained breakdown allows attention computation to proceed concurrently across both GPU groups in DeDiVAE, thereby improving GPU utilization.
The main contributions of this paper are summarized as follows:
\begin{itemize}
    \item We analyze the computation and communication patterns of Ulysses and propose an optimized version by pipelining communication and computation tasks in the denoising stage. 
    \item To tackle GPU memory limitations during video generation and inefficiency caused by offloading, we propose a module-level pipeline parallelism that separates diffusion denoising and VAE decoding across different GPUs, significantly reducing peak memory consumption and improving the generation efficiency.
    \item To enhance GPU utilization in the decoupled setup, we introduce a fine-grained decoupling strategy that further decouples DiTs into its linear-layer and attention-computation components, allowing attention operations to be distributed across all GPUs.
    \item Building upon OpenSoraPlan~\cite{pku-yuangroup2025opensora} and HunyuanVideo~\cite{tencent2025hunyuanvideo}, we evaluate PipeDiT by measuring single-timestep runtime and multi-prompt inference latency under various configurations on two 8-GPU systems. The experimental results demonstrate the effectiveness and scalability of our optimizations.
\end{itemize}

\section{Preliminaries \& Motivations}

\begin{figure}[t]
\centering
\includegraphics[width=1.\columnwidth]{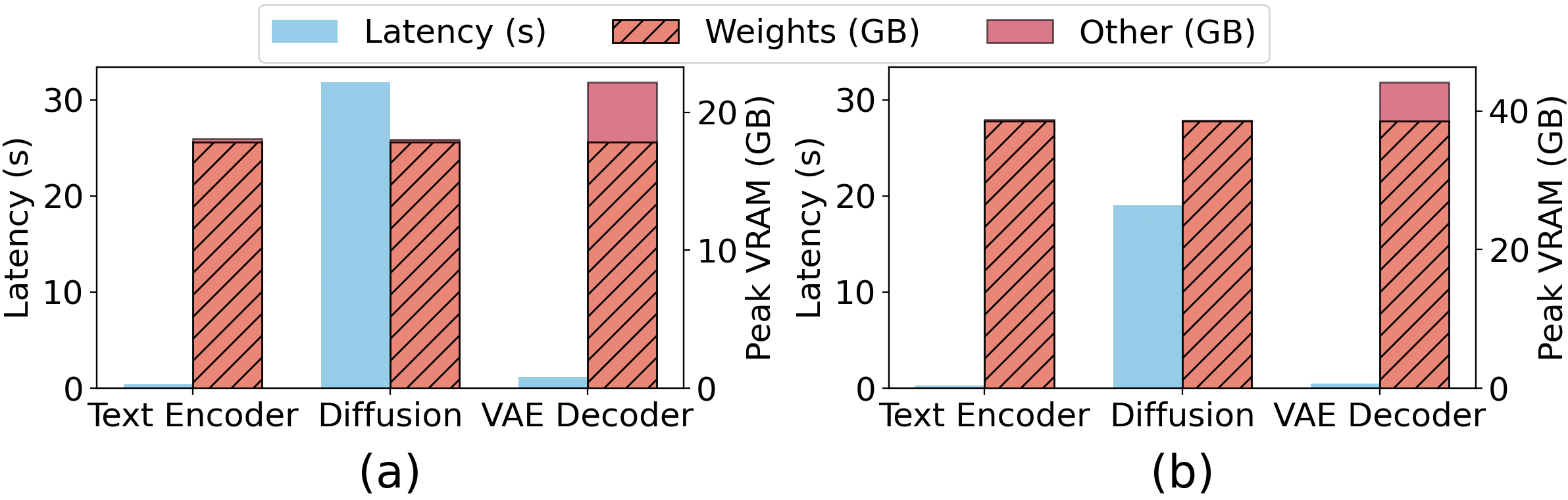} 
\caption{Latency and peak GPU memory usage of each component during inference (using eight GPUs with SP) for a single prompt in (a) OpenSoraPlan~\cite{pku-yuangroup2025opensora} model with a resolution of 480×352×65 and 50 timesteps (b) HunyuanVideo~\cite{kong2024hunyuanvideo} model with a resolution of 256×128×33 and 50 timesteps.}
\label{fig:latencybreakdown}
\end{figure}

\textbf{Diffusion-based video generation} comprises two stages as shown in Fig.~\ref{fig:dit-process}: a Diffusion-based denoising stage that refines a latent representation over multiple timesteps and a variational autoencoder (VAE)-based decoding stage that upsamples the latent into a full-resolution video. The denoising stage is computationally heavy due to attention operations, while the decoding stage is highly memory-intensive due to upsampling to target resolution and frame rate. We conduct a preliminary benchmark using two state-of-the-art video generation frameworks, OpenSoraPlan~\cite{pku-yuangroup2025opensora} and HunyuanVideo~\cite{tencent2025hunyuanvideo} as shown in Fig.~\ref{fig:latencybreakdown}. This indicates that the diffusion stage takes significantly longer than the other two stages, but its memory usage is relatively small. Without offloading enabled, the VAE Decoder peaks at 44GB of memory when decoding the 256×128×33 latent, which is largely because the model parameters occupy a substantial amount of GPU memory. It is evident that during the entire diffusion-based video generation process, the denoising of the latent becomes the time bottleneck, while the decoding of the latent is the memory bottleneck. 

In current mainstream video generation models, the diffusion backbone and the VAE decoder are typically colocated, which leads to serialized execution of diffusion computation and VAE upsampling. Without employing offloading or other memory-saving techniques, this design results in significant additional and inefficient memory consumption. Therefore, colocating the diffusion model and the VAE decoder is unfavorable for parallel video generation and hinders the generation of higher-resolution videos.
The experimental results show that under the single-GPU memory constraint of 48 GB, OpenSoraPlan is unable to generate videos with resolutions larger than 1024×576×97 without offloading. Due to its larger model weights, HunyuanVideo cannot generate videos beyond 256×128×33 in resolution (see the experimental section for details).


\textbf{Offloading} is a commonly used strategy to reduce GPU memory consumption during inference~\cite{abul2025diffusion,chen2024opportunities}. This strategy saves GPU memory by dynamically transferring model weights between the CPU and GPU. The primary advantage of this strategy is its implementation simplicity and its effectiveness in enabling higher-resolution video generation with limited GPU memory. Accordingly, offloading is adopted by several large-scale video generation systems—such as HunyuanVideo~\cite{tencent2025hunyuanvideo}, Wan~\cite{wan2025wan}, and OpenSoraPlan~\cite{pku-yuangroup2025opensora}.
\textit{However, offloading introduces significant CPU-GPU data transfer overhead, which depends on model size and bandwidth.} The offloading overhead may easily dominate and hurt efficiency, while it cannot run the inference without offloading due to the high memory consumption of video generation.

\begin{figure}[t]
\centering
\includegraphics[width=1\columnwidth]{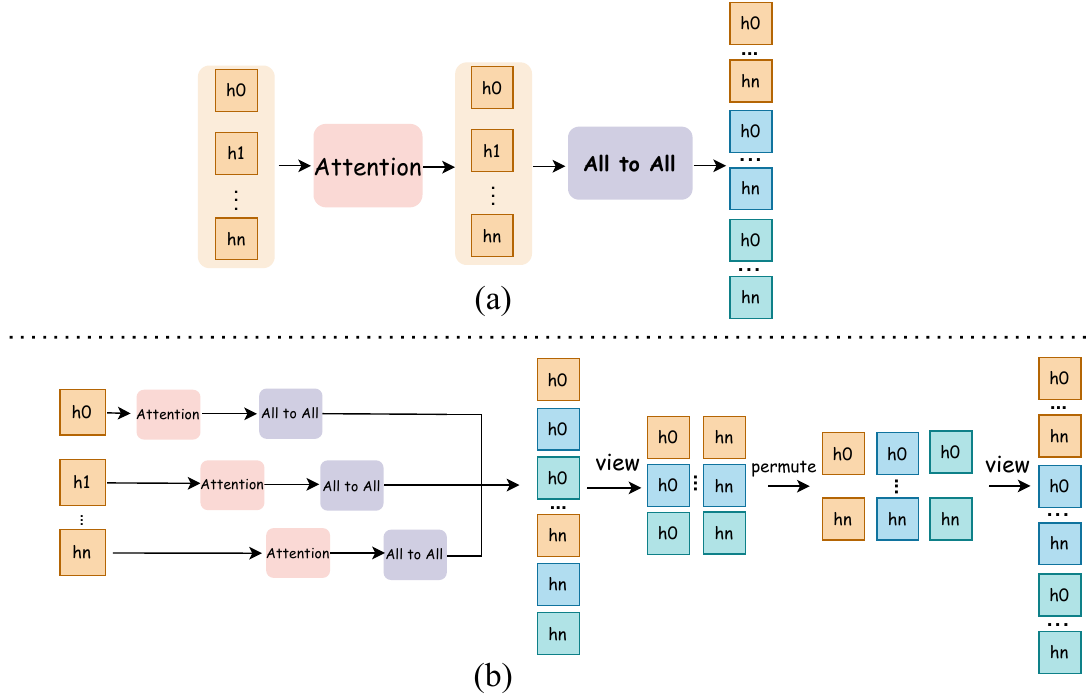} 
\caption{(a) The execution process of Ulysses, where computation and communication are executed sequentially. (b) Our optimized SP (PipeSP) by pipelining communication and computation. The subsequent post-processing resolves the misalignment issue introduced by the pipelining.}
\label{fig:sp}
\end{figure}

\textbf{Sequence parallelism (SP)} \cite{li2023lightseq,wang2025flexsp,wu2024loongserve} like Ulysses is a technique used to accelerate the processing of long input sequences on multiple GPUs. Ulysses achieves parallel computation by splitting along the attention head dimension, with different GPUs processing different attention heads. 
The computation model of Ulysses is illustrated in Fig.~\ref{fig:sp}(a). After each GPU computes its portion of the sub-sequence’s $\mathbf{Q}$, $\mathbf{K}$, and $\mathbf{V}$, three rounds of All-to-All communication are used to distribute $\mathbf{Q}$, $\mathbf{K}$, and $\mathbf{V}$ along the attention head dimension. The GPUs then concatenate the gathered $\mathbf{Q}$, $\mathbf{K}$, and $\mathbf{V}$ to form a complete sequence, but with only partial attention heads. After all attention heads have been computed, a final round of All-to-All communication is used to disperse the results along the sequence dimension and collect the data along the attention head dimension, resulting in a hidden state with partial sequence length but full attention heads.
However, in the original Ulysses kernel, as illustrated in Fig.~\ref{fig:sp}(a), a single All-to-All operation is issued only after all attention heads have been computed; \textit{during the waiting for this communication, GPUs remain idle, resulting in a waste of computational resources.}

\section{Methodology}

\begin{algorithm}[!t]

\caption{PipeSP: Overlapping Computation and Communication in SP}

\begin{algorithmic}[1]

\REQUIRE $\mathit{Q} \in \mathbb{R}^{B\times h\times S\times D}$, \ $\mathit{K}$, $\mathit{V}$,  $\mathit{attention\_mask}$

\STATE Initialize $\mathit{chunks}$, $\mathit{results}$, $\mathit{event\_lst}$

\FOR{$j \gets 0$ to $h - 1$}
    \STATE $\mathit{result} \gets \texttt{attention}(\mathit{Q}[:,j,:,:], \mathit{K}[:,j,:,:], \mathit{V}[:,j,:,:], \mathit{attention\_mask}[:,j,:,:])$

    \STATE Append $\mathit{result}$ to $\mathit{results}$
    \STATE Record event $\mathit{event\_lst}[j]$
    \STATE Wait on CUDA stream for $\mathit{event\_lst}[j]$
    \STATE $\mathit{hidden\_states} \gets \texttt{All\_to\_All}(\mathit{results}[j])$
    \STATE Append $\mathit{hidden\_states}$ to $\mathit{chunks}$
\ENDFOR
\STATE $\mathit{hidden\_states} \gets \texttt{concat}(\mathit{chunks}, \text{dim}=1)$
\STATE $\mathit{hidden\_states} \gets \texttt{view}(-1, \mathit{h}, \mathit{n}, \mathit{D})$
\STATE $\mathit{hidden\_states} \gets \texttt{permute}(0, 2, 1, 3)$
\STATE $\mathit{hidden\_states} \gets \texttt{view}(-1, \mathit{h}\times\mathit{n}, \mathit{D})$

\end{algorithmic}\label{algo:pipesp}

\end{algorithm}

\subsection{Pipelining Computation and Communication in SP}
To address the issue of serial communication and computation in Ulysses, we propose a pipelined SP (PipeSP) algorithm that partitions the computation of attention along the head dimension and issues an All-to-All immediately after each head is processed, as illustrated in Fig.~\ref{fig:sp}(b), thus overlapping communication with computation to improve the computation efficiency. Specifically, in the attention layer that has $n$ heads, each head is processed independently. Thus, the $n$ heads can be partitioned into $n$ independent attention operations, each of which has only one head. After each head has been computed at its attention, its result can be communicated with other GPUs via an All-to-All operation. Thus, the operations of attention (computation) and All-to-All (communication) form a pipeline, which keeps GPU resources be fully utilized during the inference process. After the results of all heads have been gathered, a layout transformation is performed to align the result be identical with that without pipelining.

The PipeSP algorithm is shown in Algorithm~\ref{algo:pipesp}. In lines 1–9, the $\mathbf{Q}$, $\mathbf{K}$, and $\mathbf{V}$ tensors are partitioned along the attention head dimension, and for each head, attention is computed over the full sequence. An event is recorded to mark the completion of this computation, and once all GPUs have completed the corresponding event, an All-to-All is triggered. In this step, each GPU receives the portion of the attention output corresponding to its local sequence slice for a single head. Lines 10–13 perform post-processing to reorder the collected results. This reordering is necessary because the optimized method collects attention outputs one head at a time, whereas the original method gathered them in a total different way, resulting in a misalignment shown in Fig.~\ref{fig:sp}(b). 
To resolve this, the tensor must be reshaped and permuted using the sequence \texttt{view} $\rightarrow$ \texttt{permute} $\rightarrow$ \texttt{view}. Specifically, \texttt{view(-1, h, n, D)} reshapes the head dimension into a 2D layout of $[h, n]$, \texttt{permute(0, 2, 1, 3)} swaps the GPU and head axes, and the final \texttt{view(-1, nh, D)} restores the expected layout. Mathematically, this process ensures the final tensor matches the original layout expected by the attention module, while enabling efficient communication–computation overlap. A formal proof of the correctness of the \texttt{view}--\texttt{permute}--\texttt{view} transformation is provided in the Supplementary Material.

\begin{figure*}[!t]
\centering
\includegraphics[width=0.9\textwidth]{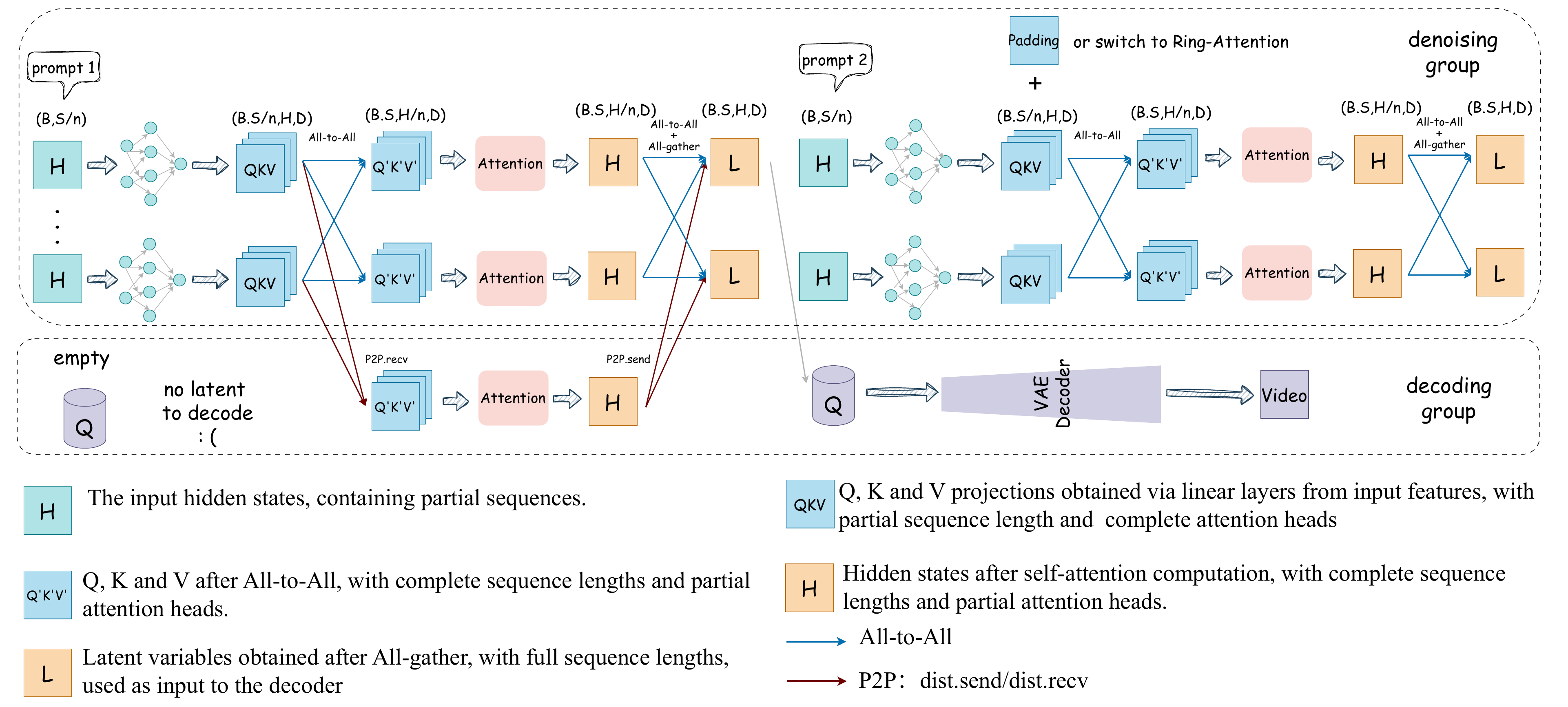} 
\caption{In the prompt 1 stage, the Denoising\,GPUs transmit the computed \textbf{Q}, \textbf{K}, and \textbf{V} tensors to the Decoding\,GPUs, enabling parallel attention computation across both groups.  In the prompt 2 stage, the Denoising\,GPUs perform attention computation independently, while the Decoding\,GPUs execute decoding in parallel.}
\label{fig:dedivae}
\end{figure*}

\subsection{Memory-Efficient Diffusion–VAE Decoupling}
To address the issues of low computational efficiency and poor GPU memory utilization caused by colocating the diffusion model and the VAE decoder, we propose \textbf{Diffusion–VAE Module Decoupling} (DeDiVAE) by breaking down the Diffusion module and the VAE module to two
disjoint GPU groups:
\emph{Denoising\,Group} and \emph{Decoding\,Group}. Specifically, for a given $N$-GPU system for video generation with DiT, DeDiVAE splits the $N$ GPUs to $N_\mathrm{denoise}$ GPUs as the \emph{Denoising\,Group} and the other $N_{\mathrm{decode}}=N-N_\mathrm{denoise}$ GPUs as the \emph{Decoding\,Group}. Accordingly, full video generation model is split into the Diffusion backbone stored in the \emph{Denoising\,Group}, and the VAE decoder stored in the \emph{Decoding\,Group}. The decoupling effectively avoids the OOM problem for large models and high-resolution video generation. The latent outputs of the \emph{Denoising\,Group} should be sent the \emph{Denoising\,Group} to generate the video. 

At first glance, DeDiVAE might also lead to idle periods due to the data dependency between the two groups, potentially affecting inference efficiency. This issue can be addressed by implementing a pipeline execution with multiple prompts. Given that a video generation service typically handles multiple ongoing queries, multiple prompts (or queries) can be pipelined within our decoupled structure as demonstrated in Fig.~\ref{fig:dedivae}. The decoding execution with VAE of the first prompt can be overlapped with the denoising execution with diffusion of the second prompt, which allows both GPU groups to keep busy. To maximize the utilization of GPU resources, we provide an effective analysis on how many GPUs should be assigned to the two groups.

\textbf{Optimal GPU partitioning.}  
Given $N$ GPUs for inference, there are $N_{\mathrm{denoise}}$ GPUs in the \emph{Denoising\,Group} and $N_{\mathrm{decode}}$ GPUs in the \emph{Decoding\,Group}. Let $T_{\mathrm{denoise}}$ denote the time of denoising one prompt on a single GPU and $T_{\mathrm{decode}}$ denote the time of decoding one latent on a single GPU.
During inference we accelerate the denoising with SP across $N_{\mathrm{denoise}}$ GPUs, while the decoder uses data parallelism across $N_{\mathrm{decode}}$ GPUs. Since the total workloads of inference are unchanged, to enable an maximal overlap is to make the execution time of the groups be identical. Thus, the first-order balance condition is

\begin{equation*}
\begin{aligned}
( \frac{T_{\text{denoise}}}{N_{\text{denoise}}} + T_{\text{comm}}) N_{\text{decode}}
&\approx T_{\mathrm{decode}}, \\[6pt]
\Longrightarrow\quad
N_{\mathrm{decode}}
&\approx 
\left\lceil 
\frac{T_{\mathrm{decode}}}{T_{\mathrm{decode}} +  T_{\mathrm{denoise}}} \, N
\right\rceil,
\end{aligned}
\end{equation*}
which yields the optimal $N_{\mathrm{decode}}$ that maximizes GPU
utilization: both stages finish a micro-batch in approximately the same
time, preventing either group of GPUs from idling while the other is
still computing. Since intra-node GPU communication is very fast, and PipeSP overlaps communication with computation to hide most of the communication overhead, omitting $T_{\mathrm{comm}}$ does not affect the resulting resource allocation.

In practice, though we assign the GPUs in a balance way, the execution time of the diffusion stage may still dominate the overall execution time. To improve the efficiency, we design a new co-processing approach in DeDiVAE as introduced in the following section.

\subsection{Attention Co-processing}
When denoising process is much slower than a VAE decoding, 
the Decoding\,GPUs idle during most of the generation window, and the pipeline cannot achieve a full overlap.
Therefore, we propose \textbf{Attention Co-processing} (Aco) to utilize the idle time of the \emph{Decoding\,Group}. We further split the DiT block into two disjoint kernels:
\begin{itemize}\setlength{\itemsep}{2pt}
  \item \emph{Linear projections}: $\mathbf Q=\mathbf XW_Q,\;
        \mathbf K=\mathbf XW_K,\;
        \mathbf V=\mathbf XW_V$,
  \item \emph{Attention kernel}: $\operatorname{Attn}(\mathbf Q,\mathbf K,\mathbf V)$,
\end{itemize}
and assign them to the two GPU groups.
The Denoising\,GPUs keep the DiT weights and compute the
linear projections; immediately afterwards they transmit the resulting
$\mathbf Q,\mathbf K,\mathbf V$ tensors via point-to-point links to the
Decoding\,GPUs, when Decoding\,GPUs are not decoding latents.  
Because the attention kernel depends only on
$\mathbf Q,\mathbf K,\mathbf V$, and the computations of different attention heads are independent in multi-head attention, the
Decoding\,GPUs can execute it autonomously.

As shown in Fig.~\ref{fig:dedivae}, in the prompt 1 stage, since there is no latent requiring for decoding, the idle Decoding\,GPUs can be leveraged to assist with attention computation. Specifically, the Denoising\,GPUs first perform intra-group All-to-All communication, followed by P2P communication to send the $\mathbf Q,\mathbf K,\mathbf V$ tensors to the Decoding\,GPUs. Both groups then execute attention computations in parallel. Afterward, the Denoising\,GPUs aggregate the results via intra-group All-to-All and inter-group P2P communication, obtaining latents with partial sequence length but complete attention heads. Afterwards, the Denoising\,GPUs push the latent outputs into a shared queue that is accessible to all processes. The Decoding\,GPUs then retrieve them from the queue for subsequent decoding.

In the prompt 2 stage, as the decoding queue becomes non-empty, the Decoding\,GPUs are occupied with decoding latents from the queue. Consequently, the Denoising\,GPUs must perform attention computation independently. During this stage, denoising and decoding proceed in parallel. It is worth noting that if the number of attention heads is not divisible by the number of Denoising GPUs, there are two possible strategies to handle this. For models that only adopt Ulysses, such as OpenSoraPlan~\cite{pku-yuangroup2025opensora}, head dimension padding must be introduced to ensure balanced workload distribution. In contrast, models like HunyuanVideo~\cite{tencent2025hunyuanvideo} and Wan~\cite{wan2025wan} adopt Unified Sequence Parallelism (USP)~\cite{fang2024usp}, which allows flexible configuration of both the Ulysses degree and the Ring-Attention degree. When the number of heads is not divisible, we can switch Denoising\,GPUs from Ulysses to Ring-Attention, effectively changing the parallelism from head-wise splitting to sequence-wise splitting, thereby avoiding the overhead of padding and improving GPU utilization.

\textbf{Performance Analysis.} Let $t_L$ and $t_A$ denote the wall-clock time of one linear-projection
and one attention kernel on Denoising\,GPUs, respectively, and let
$N_{\mathrm{denoise}}$ and $N_{\mathrm{decode}}$ be the two GPU group sizes
($N_{\mathrm{denoise}}+N_{\mathrm{decode}}=N$).  
In the baseline decoupling only the denoising group participates:
\begin{equation}  
T_{\mathrm{baseline}}
  = t_L + t_A .
\end{equation}
With Aco, the linear part still costs $t_L$, but the
attention time scales inversely with the total number of GPUs
that now share the work:
\begin{equation}
T_{\mathrm{coop}}
  = t_L \;+\;
    t_A\,\frac{N_{\mathrm{denoise}}}{N_{\mathrm{denoise}}+N_{\mathrm{decode}}}.
\end{equation}
Hence the theoretical speed-up is
\begin{equation}
  S = \frac{T_{\mathrm{baseline}}}{T_{\mathrm{coop}}}
  = \frac{t_L+t_A}
         {t_L+t_A\frac{N_{\mathrm{denoise}}}{N}}.
\end{equation}
Note that the above analysis assumes the number of attention heads $H$ is divisible by the number of Denoising\,GPUs $N_{\mathrm{denoise}}$.
If not divisible, and switching between Ulysses and Ring-Attention as in HunyuanVideo is not supported, then padding is required to balance the workload, leading to wasted GPU resources.  
For example, if $H = 24$ and 7 GPUs are used for denoising, padding increases the head count to 28 so that each GPU handles 4 heads. However, only 6 GPUs are effectively needed, and one GPU performs redundant computations.  
Our Attention Co-processing solves this issue by avoiding padding and ensuring all GPUs perform meaningful work, even when $H$ is not divisible by $N_{\mathrm{denoise}}$.

\begin{table*}[ht]
\setlength{\tabcolsep}{1.6pt}
\begin{tabular}{*{10}{c}|*{12}{c}}
  \toprule
  \multirow{3}*{Resolution}&\multicolumn{9}{c|}{\textbf{OpenSoraPlan (A6000)}}&\multicolumn{9}{c}{\textbf{OpenSoraPlan (L40)}}\\
  \cmidrule(lr){2-20}
  
  & \multicolumn{3}{c}{10} & \multicolumn{3}{c}{30}& \multicolumn{3}{c|}{50}&\multicolumn{3}{c}{10}  & \multicolumn{3}{c}{30}&\multicolumn{3}{c}{50}\\
  \cmidrule(lr){2-4}\cmidrule(lr){5-7}\cmidrule(lr){8-10}\cmidrule(lr){11-13}\cmidrule(lr){14-16}\cmidrule(lr){17-19}
  &base&opt&spd$\uparrow$&base&opt&spd$\uparrow$&base&opt&spd$\uparrow$&base&opt&spd$\uparrow$&base&opt&spd$\uparrow$&base&opt&spd$\uparrow$
\\
  \midrule

$480\times352\times97$ &227 &107  &2.12×   &420 &304 &1.38×  &622 &502  &1.24×&252 &154 &1.64×  &492  &\textbf{407} &1.21×   &738 &\textbf{657} &1.12× &\\

$640\times352\times97$ &257  &135  &1.90×   &522  &389 &1.34× &786 &643 &1.22× &303  &\textbf{206} &1.47× &650  &\textbf{545} &1.19× &983 &\textbf{883} &1.11× &\\

$800\times592\times97$ &520  &397  &1.31×  &1257  &\textbf{1097} &1.15×  &1994 &\textbf{1766} &1.13×&646  &\textbf{517} &1.25× &1609 &\textbf{1441} &1.12× &2570 &\textbf{2373} &1.08× &\\

$1024\times576\times97$ &555  &430  &1.29×  &1360  &\textbf{1144} &1.19× &2162 &\textbf{1832} &1.18×&731  &\textbf{591} &1.24× &1836 &\textbf{1639} &1.12×  &2940 &\textbf{2689} &1.09× &\\

  \toprule
  \multirow{3}*{Resolution}&\multicolumn{9}{c|}{\textbf{ HunyuanVideo (A6000)}}&\multicolumn{9}{c}{\textbf{HunyuanVideo (L40})}\\
    \cmidrule(lr){2-20}
  
  & \multicolumn{3}{c}{10} & \multicolumn{3}{c}{30}& \multicolumn{3}{c|}{50}&\multicolumn{3}{c}{10} & \multicolumn{3}{c}{30}& \multicolumn{3}{c}{50}\\
  \cmidrule(lr){2-4}\cmidrule(lr){5-7}\cmidrule(lr){8-10}\cmidrule(lr){11-13}\cmidrule(lr){14-16}\cmidrule(lr){17-19}
  &base&opt&spd$\uparrow$&base&opt&spd$\uparrow$&base&opt&spd$\uparrow$&base&opt&spd$\uparrow$&base&opt&spd$\uparrow$&base&opt&spd$\uparrow$&
\\
  \midrule
  
$480\times352\times97$ & 540 & \textbf{165} & 3.27× & 767 & \textbf{445} & 1.72× & 965 & \textbf{726} & 1.33×& 676 & \textbf{229} & 2.95× & 992 & \textbf{649} & 1.53×  & 1350 & \textbf{1068} & 1.26× \\

$640\times352\times97$ & 593 & \textbf{191} & 3.10×  & 865 & \textbf{531} & 1.63× & 1142 & \textbf{907} & 1.26×& 760 & \textbf{295} & 2.58× & 1231 & \textbf{843} & 1.46×  & 1702 & \textbf{1392} & 1.22× \\

$800\times592\times97$ & 1082 & \textbf{506} & 2.14×  & 1880 & \textbf{1492} & 1.26×  & 2686 & \textbf{2470} & 1.09×& 1694 & \textbf{923} & 1.84×  & 3291 & \textbf{2702} & 1.22×  & 4898 & \textbf{4482} & 1.09× \\

$1024\times576\times97$ & 1399 & \textbf{729} & 1.92×  & 2545 & \textbf{2090} & 1.22× & 3726 & \textbf{3453} & 1.08×  & 2237 & \textbf{1333} & 1.68× & 4576 & \textbf{3894} & 1.18× & 6952 & \textbf{6453} & 1.08× \\

  \bottomrule

\end{tabular}

\caption{Latency and speedup for generating 10 videos with the baseline system and our optimized PipeDiT. Bold numbers indicate the results obtained with PipeDiT w/ Aco.}
\label{table:endtoend}
\end{table*}

\begin{table*}[!t]
\centering
\setlength{\tabcolsep}{1.3pt}
\begin{tabular}{ccccccccc}
\toprule  
\multicolumn{9}{c}{\textbf{OpenSoraPlan (A6000)}} \\
\midrule
     & 480x352x65& 480x352x129 & 640x352x65& 640x352x129 & 800x592x65 & 800x592x129& 1024x576x65        & 1024x576x129  \\
\midrule  
\rowcolor{gray!15}\noalign{\vskip -3.pt}
Baseline (s)  & 1.67  & 1.30  & 1.21 &2.10& 2.21  & 4.98  &2.57  &6.86  \\   
PipeSP(s)  & 1.73  & 1.20  & 1.69&1.83  & 2.05  & 4.74  &2.41 &6.54 \\
Speedup  & $0.97\times$  & $1.08\times$ & $0.72\times$ & $1.15\times$ & $1.08\times$  & $1.05\times$ &  $1.07\times$ & $1.05\times$ \\
\midrule  
\multicolumn{9}{c}{\textbf{OpenSoraPlan (L40)}}\\
\midrule
     & 480x352x65& 480x352x129 & 640x352x65& 640x352x129 & 800x592x65 & 800x592x129& 1024x576x65        & 1024x576x129  \\
\midrule  
\rowcolor{gray!15}\noalign{\vskip -3.pt}
Baseline (s)  & 1.36  & 1.66  &1.05  &2.44& 2.87  &7.34   &3.61  &10.30  \\   
PipeSP (s)  & 1.42  & 1.57  &1.01 &2.34  & 2.74  &7.05   &3.44 &9.95 \\
Speedup  & 0.96$\times$  & 1.06$\times$ &1.04$\times$&1.04$\times$& 1.05$\times$  &1.04$\times$ &  1.05$\times$&1.04$\times$ \\
\bottomrule

\end{tabular}

\caption{Improvement in per-timestep latency with our PipeSP.}
\label{table:pipesp}
\end{table*}

\section{Evaluation}
\subsection{Experimental Setups}
\textbf{Baselines.}
We implement our PipeDiT on two state-of-the-art open-source video generation systems OpenSoraPlan at v1.3.0\footnote{\url{https://github.com/PKU-YuanGroup/Open-Sora-Plan}} and HunyuanVideo\footnote{\url{https://github.com/Tencent-Hunyuan/HunyuanVideo}}. As the generation algorithm of PipeDiT is identical with OpenSoraPlan and HunyuanVideo, the generated videos are identical, so we mainly compare the time and memory efficiency. Note that OpenSoraPlan uses Ulysses, while HunyuanVideo integrates USP in xDiT~\cite{fang2024xdit}. The model sizes for OpenSoraPlan and HunyuanVideo are 2B and 13B parameters, respectively.

\noindent
\textbf{Performance Metrics.}
The performance metrics are video generation efficiency and GPU memory consumption. The efficiency metrics consist of two aspects. The first is the latency per timestep, which measures the optimization effect of PipeSP. The second metric is the overall latency for generating multiple videos from consecutive prompts, which measures the overall optimization effect of PipeDiT.

\noindent
\textbf{Testbeds.}
All experiments are conducted on two 8-GPU systems: 1) eight NVIDIA  RTX A6000 48GB GPUs and 2) eight NVIDIA L40 48GB GPUs. More environment information can be found in Supplementary Material. 


\subsection{End-to-End Performance}
We configure different video resolutions, commonly used diffusion timesteps (10, 30, and 50), and 10 prompts\footnote{Due the page limit, we put the results with more comprehensive configurations in Supplementary Material.} to compare generation latency as shown in Table~\ref{table:endtoend}, where ``base'' indicates the baseline using offloading, ``opt'' indicates our PipeDiT, and ``spd'' indicates the speedup of PipeDiT over the baseline. 
Since Aco does not always achieve improvement especially on low workload video generation, bold numbers indicate the results generated using PipeDiT w/ Aco.



From the results in Table~\ref{table:endtoend}, our PipeDiT always achieves faster inference speed by $1.08\times$-$3.27\times$ than baseline both OpenSoraPlan and HunyuanVideo. Particularly, PipeDiT yields the most notable speedups under lower resolutions, fewer frames, and shorter timesteps, reaching up to $3.27\times$ (up to $4.02\times$ as shown in Supplementary Material). As the resolution, frame count, and timesteps increase, the benefit diminishes since the computation time dominates and the relative impact of data transfer decreases, making offloading less of a bottleneck.
Despite this, our PipeDiT on OpenSoraPlan with the A6000 platform still achieves $1.18\times$ improvement over the baseline in the highest setting.

For different models, HunyuanVideo has more parameters, so its offloading takes longer time, making PipeDiT more effective to HunyuanVideo under lower resolutions and timesteps. In contrast, under higher resolutions and timesteps, the optimizations of PipeDiT in HunyuanVideo bring greater gains to OpenSoraPlan due to its shorter computation time. For different hardware, because the A6000 GPUs are connected with NVLink which delivers higher communication speed than L40. Thus, PipeDiT yields shorter per-timestep computation times compared to L40, and thus it has higher improvement on A6000 than L40.

\begin{table*}[!t]
\setlength{\tabcolsep}{1.pt}
\begin{tabular}{*{22}{c}}
  \toprule
  \multicolumn{21}{c}{\textbf{OpenSoraPlan (A6000)}}\\
  \cmidrule{6-21}
    \multirow{2}*{A}&\multirow{2}*{B}&\multirow{2}*{C}&\multirow{2}*{D} & \multicolumn{3}{c}{480×352×65} & \multicolumn{2}{c}{480×352×129} & \multicolumn{2}{c}{640×352×65}& \multicolumn{2}{c}{640×352×129}& \multicolumn{2}{c}{800×592×65}& \multicolumn{2}{c}{800×592×129}& \multicolumn{2}{c}{  1024×576×65}& \multicolumn{2}{c}{  1024×576×129}\\
  \cmidrule(lr){6-7}\cmidrule(lr){8-9}\cmidrule(lr){10-11}\cmidrule(lr){12-13}\cmidrule(lr){14-15}\cmidrule(lr){16-17}\cmidrule(lr){18-19}\cmidrule(lr){20-21}
  & & & & &T(s)$\downarrow$&Spd$\uparrow$&T(s)$\downarrow$&Spd$\uparrow$&T(s)$\downarrow$&Spd$\uparrow$&T(s)$\downarrow$&Spd$\uparrow$&T(s)$\downarrow$&Spd$\uparrow$&T(s)$\downarrow$&Spd$\uparrow$&T(s)$\downarrow$&Spd$\uparrow$ \\
  \midrule
  \rowcolor{gray!15}\noalign{\vskip -3.pt}
\ding{51}&\ding{55}&\ding{55}&\ding{55}& &314&1×&529&1×&368&1×&665&1×&777&1×&1875&1×&851&1×&1995&1×\\

\ding{55}&\ding{51}&\ding{55}&\ding{55}&&217&1.45×&452&1.17×&234&1.57×&500&1.33×&649&1.20×&1872&1.00×&702&1.21×&2138&0.93×\\
\ding{55}&\ding{51}&\ding{51}&\ding{55}&&200&1.57×&390&1.36×&250&1.47×&509&1.31×&649&1.20×&1847&1.02×&717&1.19×&1936&1.03×\\
\ding{55}&\ding{51}&\ding{51}&\ding{51}&&261&1.20×&414&1.28×&296&1.24×&507&1.31×&645&1.20×&1652&1.14×&683&1.25×&1690&1.18×\\

  \midrule
  
\multicolumn{21}{c}{\textbf{HunyuanVideo (A6000)}}\\
  \midrule
  \rowcolor{gray!15}\noalign{\vskip -3.pt}
  \ding{51}&\ding{55}&\ding{55}&\ding{55}& &636&1×&911&1×&695&1×&1104&1×&1294&1×&2681&1×&1676&1×&3733&1×\\
\ding{55}&\ding{51}&\ding{55}&\ding{55}&&340&1.87×&681&1.34×&403&1.72×&824&1.34×&984&1.32×&2501&1.07×&1374&1.22×&3680&1.01×\\
\ding{55}&\ding{51}&\ding{51}&\ding{55}&&345&1.84×&701&1.30×&404&1.72×&824&1.34×&983&1.32×&2499&1.07×&1374&1.22×&3675&1.02×\\
\ding{55}&\ding{51}&\ding{51}&\ding{51}&&327&1.94×&595&1.53×&396&1.76×&741&1.49×&942&1.37×&2259&1.19×&1242&1.35×&3090&1.21×\\

\bottomrule

\end{tabular}

\caption{Efficiency improvement of different optimization methods.}
\label{table:ablation}
\end{table*}

\subsection{Effectiveness of PipeSP}
To evaluate the effectiveness of PipeSP, we use eight representative resolution and frame configurations and measure the per-timestep latency, as shown in Table~\ref{table:pipesp}. Notably, PipeSP achieves a 15\% performance improvement under the 640×352×129 configuration. The results also indicate that the optimization achieves the best performance under moderate resolutions, as it strikes a balance between overly short and excessively long computation times. When the resolution is low, the computation time is short, and the communication overhead introduced by overlap can offset its benefits. Conversely, at high resolutions, the proportion of communication time becomes less significant relative to the overall computation time, thereby reducing the optimization gains.

\begin{table}[!t]
\centering
\resizebox{1\columnwidth}{!}{
\begin{tabular}{ccccccccc}
\toprule
\multirow{3}*{Methods}&\multicolumn{8}{c}{\textbf{OpenSoraPlan}}\\
\cmidrule{2-9}
     & \multicolumn{2}{c}{480x352x129}& \multicolumn{2}{c}{640x352x129}& \multicolumn{2}{c}{800x592x129}&\multicolumn{2}{c}{1024x576x129}\\
     \cmidrule(lr){2-3}\cmidrule(lr){4-5}\cmidrule(lr){6-7}\cmidrule(lr){8-9}
     &Mem &$\downarrow$\%&Mem &$\downarrow$\%&Mem&$\downarrow$\%&Mem &$\downarrow$\% \\
\midrule
\rowcolor{gray!15}\noalign{\vskip -3.pt}
\textbf{Baseline}     & 26.5   & --     & 29.4   & --     & 39.8   & --     & OOM   & --     \\
Offloading            & 18.4   & 30.6\% & 18.4   & 37.4\% & 19.1   & 52.0\% & 28.3  & 41.0\% \\
DeDiVAE            & 18.0   & 32.1\% & 18.2   & 38.1\% & 18.6   & 53.3\% & 28.1  & 41.5\% \\
\midrule
& \multicolumn{8}{c}{\textbf{HunyuanVideo}} \\
\midrule
\rowcolor{gray!15}\noalign{\vskip -3.pt}
\textbf{Baseline}     & OOM   & --     & OOM   & --     & OOM   & --     & OOM   & --     \\
Offloading            & 29.37 & 38.8\% & 29.38 & 38.8\% & 32.97 & 31.3\% & 33.01 & 31.2\% \\
DeDiVAE            & 41.44 & 13.6\% & 41.43 & 13.7\% & 41.45 & 13.6\% & 42.12 & 12.2\% \\
\bottomrule

\end{tabular}
}
\caption{Peak GPU memory usage (GB) and reduction ratio.}
\label{table:memory}
\end{table}

\subsection{Memory Efficiency of DeDiVAE}
For the memory optimization comparison, we compare our DeDiVAE with the original implementation w/o offloading as the baseline. The second row in Table~\ref{table:memory} presents the original implementation with offloading, while the third row shows the results of our DeDiVAE approach. As shown in the table, both our method and the offloading strategy significantly reduce memory consumption during inference. The baseline implementation fails with an OOM error under the highest resolution setting, indicating that without any memory optimization strategies, the model is unable to generate videos beyond $1024\times576\times129$ in OpenSoraPlan and $480\times352\times129$ in HunyuanVideo.

It should be noted that in HunyuanVideo, our DeDiVAE demonstrates greater peak memory consumption than the offloading method. This occurs because we colocated the text encoder with the VAE decoder. Given the substantial size of the text encoder in HunyuanVideo, colocating it with the DiT module, as done in OpenSoraPlan, is not feasible. This demonstrates that our method allows adaptable management of module positioning based on the attributes of the model.

\begin{figure}[!h]
\centering
\includegraphics[width=0.9\columnwidth]{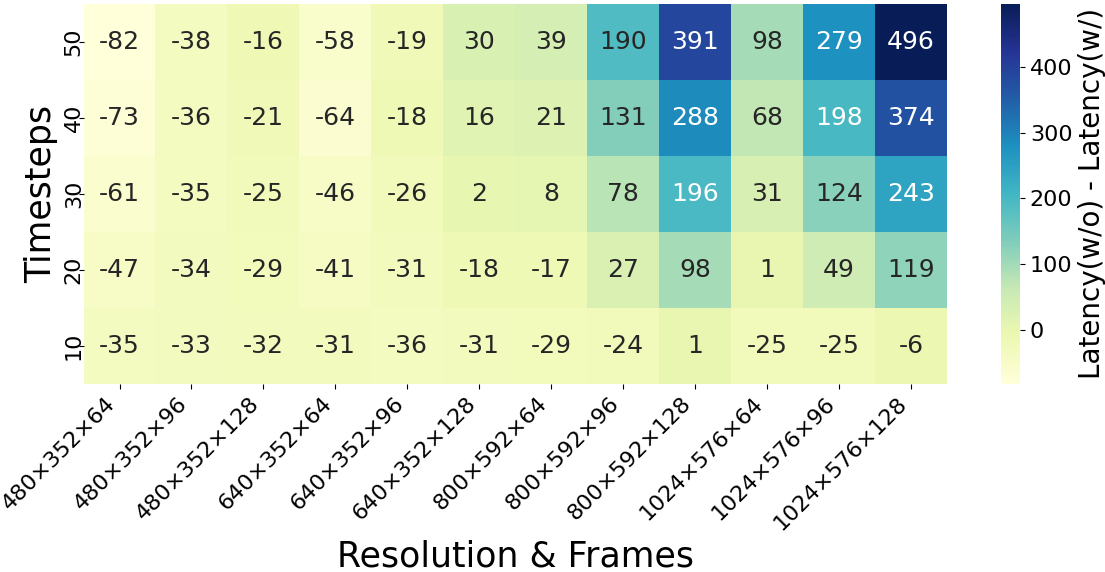} 
\caption{The heatmap of the latency difference between the two methods: (1) PipeDiT w/o Aco and (2) PipeDiT w/ Aco. 
}
\label{fig:aco}
\end{figure}

\subsection{Ablation Study}
To evaluate the performance improvements of various optimization methods, we fixed the number of timesteps to 30 and selected eight different resolutions and frame settings for ablation studies. The results are shown in Table~\ref{table:ablation}, where ``A'' indicates the baseline offloading method, ``B'' indicates DeDiVAE, ``C'' indicates PipeSP, and ``D'' refers to Aco. 

The results show that PipeSP demonstrates significant performance improvements on OpenSoraPlan. This is primarily because OpenSoraPlan incurs longer All-to-All communication times, and due to its non-modular design, we were able to partially overlap the computation of \textbf{Q}, \textbf{K}, and \textbf{V} with the three All-to-All communications that occur before the attention computation.

For DeDiVAE, it achieves substantial efficiency gains under lower resolutions. However, as the resolution increases, the drawback of having fewer GPUs allocated to denoising becomes more apparent. 
Introducing Aco helps to address this limitation. Even under the highest resolution setting, the combined approach still delivers considerable performance benefits. The performance difference of PipeDiT w/ Aco and w/o Aco is shown in Fig.~\ref{fig:aco}, which indicates that Aco improves performance on high workload tasks.

\section{Conclusion}
In this study, we proposed a system-level optimization system named PipeDiT for accelerating the video generation with diffusion transformer (DiT) based models. There are three key innovations in PipeDiT: 1) a pipelining algorithm named PipeSP for sequence parallelism (SP) for enable overlapping between communication and computation tasks, 2) a module decoupling method named DeDiVAE by breaking down the diffusion module and VAE module to two GPU groups to reduce the memory consumption, and 3) an attention co-processing approach named Aco, which leverages the idle decoding GPU group to assist with denoising module execution. Our PipeDiT is implemented atop two state-of-the-art video genrations frameworks OpenSoraPland and HunyuanVideo. Extensive experiments were conducted on two GPU systems and the results indicate that our PipeDiT not only significantly reduces memory consumption but also greatly enhances the generation efficiency.

\section*{Acknowledgments}
The research was supported in part by the National Natural Science Foundation of China (NSFC) under Grant No. 62302123, and the Shenzhen Science and Technology Program under Grant No. KJZD20240903104103005, KJZD20230923115113026, and KQTD20240729102154066.

\bibliography{aaai2026}


\clearpage
\section*{Supplementary Material}\label{sec:appendix}

\subsection{Complete End-to-End Performance Results}
In the main body, we present a subset of the End-to-End Performance Results, while Table 6 provides the complete results. We evaluate our model under multiple resolutions ranging from 480×352 to 1024×576, covering common video qualities such as 480p and 576p. The timestep values are arranged in descending order, 
where fewer timesteps lead to lower latency but also lower generation quality.

Results demonstrate that PipeDiT achieves notable acceleration across both models and platforms, proving its effectiveness for lightweight frameworks like OpenSoraPlan and large-scale systems like HunyuanVideo. PipeDiT is thus applicable to video generation models of various sizes.

Under low to medium resolutions and shorter timesteps, PipeDiT achieves up to 4.02× speedup. Even at the highest resolution and timestep, it still delivers 1.06× to
1.17× improvement. As the current trend in DiT-based inference optimization moves toward using fewer timesteps and more aggressively compressed VAE decoders to reduce denoising latency, the speedup brought by PipeDiT is expected to increase as the denoising time becomes smaller.

Additionally, some recent video generation models, such as Wan2.2, have adopted Mixture-of-Experts (MoE) architectures that significantly increase model size. In such cases, traditional offloading approaches become less viable, while PipeDiT offers a scalable and efficient alternative well-suited to these future scenarios.

\subsection{Complete Results of Ablation Study}
In the main text, we only present the ablation results on the A6000 platform. Table 7 shows the full ablation study results across all platforms, where ``A'' indicates the baseline offloading method, ``B'' indicates DeDiVAE, ``C'' indicates PipeSP, and ``D'' refers to Aco. We observe that PipeDiT achieves more significant performance improvements on the A6000 platform compared to the L40 platform. This is primarily because, under the same model and input configurations, the end-to-end latency on the A6000 is lower than that on the L40 for both the baseline and PipeDiT. As a result, the proportion of offloading time—eliminated by PipeDiT—accounts for a larger share of the total runtime on the A6000 system, leading to a higher relative speedup.

Furthermore, the A6000’s support for high-bandwidth NVLink interconnects allows more efficient inter-GPU communication compared to the PCIe-based L40. This architectural advantage further amplifies the effectiveness of PipeDiT’s PipeSP. These results demonstrate that PipeDiT can adapt well across heterogeneous hardware configurations while achieving especially notable improvements on platforms equipped with high interconnect bandwidth and lower baseline latencies.

\subsection{Consistency Proof of PipeSP}\label{app:sub:pipespproof}
As PipeSP generates misordered results which are then transformed to the original layout, we prove that the results of PipeSP is identical with the original SP.
\textbf{Resulting mis-order}  
After all heads have been processed, every GPU owns the hidden state of
its sub-sequence, but the global head order is now interleaved:
for GPU index $i\!\in\!\{0,\dots,n-1\}$ and local head
$j\!\in\!\{0,\dots,h-1\}$
\begin{equation}
k_{\mathrm{orig}}(i,j)=ih+j,\qquad
k_{\mathrm{mod}}(i,j)=j n+i.
\end{equation}

The tensor $T^{\mathrm{mod}}\!\in\!\mathbb R^{B\times H\times D}$
therefore has head $k_{\mathrm{mod}}(i,j)$ where the original order
expects $k_{\mathrm{orig}}(i,j)$; a re-ordering is mandatory.

Let $B$ be the mini-batch, $n$ the number of GPUs, $h$ heads per GPU
($H=n h$), and $D$ the head dimension. Since the sequence dimension $S$ is not involved in the operations considered in this section, it is omitted for brevity. 
Define the reshape map $\varphi_{h,n}: \mathbb{R}^{B\times H\times D}\!\longrightarrow\!\mathbb{R}^{B\times h\times n\times D}$,

\begin{equation}
(\varphi_{h,n}T)[b,j,i,d]=T[b,\,j n+i,\,d],  
\end{equation}

and its inverse
$\varphi_{h,n}^{-1}$ that merges $(i,j)$ back into a linear head index.
Let
$\pi: \mathbb R^{B\times h\times n\times D}\!\to\!
       \mathbb R^{B\times n\times h\times D},$
\begin{equation}
(\pi X)[b,i,j,d]=X[b,j,i,d].
\end{equation}

The code sequence view($-1,h,n,D$) $\to$ permute($0,2,1,3$) $\to$ view($-1,nh,D$) implements the composite map
\begin{equation}
\Psi=\varphi_{h,n}^{-1}\!\circ\!\pi\!\circ\!\varphi_{h,n}:
      \mathbb R^{B\times H\times D}\longrightarrow
      \mathbb R^{B\times H\times D}.
\end{equation}

\noindent
\textbf{Alignment proof}  
For any $b,d,i,j$ one has
\[
\begin{aligned}
(\Psi T^{\mathrm{mod}})[b,k_{\mathrm{orig}}(i,j),d]
&=(\varphi_{h,n}^{-1}\pi\varphi_{h,n}T^{\mathrm{mod}})[b,\,ih+j,\,d]\\
&=(\pi\varphi_{h,n}T^{\mathrm{mod}})[b,i,j,d]\\
&=(\varphi_{h,n}T^{\mathrm{mod}})[b,j,i,d]\\
&=T^{\mathrm{mod}}[b,\,jn+i,\,d]\\
&=T^{\mathrm{mod}}[b,k_{\mathrm{mod}}(i,j),d].
\end{aligned}
\]
Hence ${\Psi\,T^{\mathrm{mod}}=T^{\mathrm{orig}}}$:  
after the two view plus one permute, the interleaved tensor
is mapped exactly onto the head-contiguous layout expected by the
original Ulysses implementation.

\subsection{Experimental Environment}

Our experiments are conducted entirely on A6000 and L40 GPUs. The hardware configurations of the test platforms are shown in Table~\ref{hardware}.

\begin{table}[h]
\centering
\resizebox{1.0\columnwidth}{!}{
\begin{tabular}{ll}
\toprule
\textbf{Component} & \textbf{Specification} \\
\midrule
CPU    & Intel\textsuperscript{\textregistered} Xeon\textsuperscript{\textregistered} Platinum 8358, @2.60 GHz \\
GPU    & 8$\times$ Nvidia RTX A6000, 48GB GDDR6 \\
NVLink & 112.5GB/s (4$\times$) \\
PCIe   & 4.0 (x16) \\
\midrule
CPU    & Intel\textsuperscript{\textregistered} Xeon\textsuperscript{\textregistered} Platinum 8358, @2.60 GHz \\
GPU    & 8$\times$ Nvidia L40, 48GB GDDR6 \\
NVLink & - \\
PCIe   & 4.0 (x16) \\
\bottomrule
\end{tabular}
}
\caption{Testbed Hardware Specifications.}
\label{hardware}
\end{table}

\begin{center}
\centering
\resizebox{1.0\textwidth}{!}{
\begin{tabular}{*{16}{c}|*{18}{c}}
  \toprule
  \multirow{3}*{Resolution}&\multicolumn{15}{c|}{\textbf{OpenSoraPlan (A6000)}}&\multicolumn{15}{c}{\textbf{OpenSoraPlan (L40)}}\\
  \cmidrule(lr){2-32}
  
  & \multicolumn{3}{c}{10} & \multicolumn{3}{c}{20} & \multicolumn{3}{c}{30}& \multicolumn{3}{c}{40}& \multicolumn{3}{c|}{50}&\multicolumn{3}{c}{10} & \multicolumn{3}{c}{20} & \multicolumn{3}{c}{30}& \multicolumn{3}{c}{40}& \multicolumn{3}{c}{50}\\
  \cmidrule(lr){2-4}\cmidrule(lr){5-7}\cmidrule(lr){8-10}\cmidrule(lr){11-13}\cmidrule(lr){14-16}\cmidrule(lr){17-19}\cmidrule(lr){20-22}\cmidrule(lr){23-25}\cmidrule(lr){26-28}\cmidrule(lr){29-31}
  &base&opt&spd$\uparrow$&base&opt&spd$\uparrow$&base&opt&spd$\uparrow$&base&opt&spd$\uparrow$&base&opt&spd$\uparrow$&base&opt&spd$\uparrow$&base&opt&spd$\uparrow$&base&opt&spd$\uparrow$&base&opt&spd$\uparrow$&base&opt&spd$\uparrow$
\\
  \midrule
480×352×65 &177  &73  &2.42×   & 240 &138 &1.74×   &314 &200 &1.57×  &379 &264 &1.44×  &437 &329  &1.33×&204  &105  &1.94×  & 274 &202 &1.35×  &355 &\textbf{298} &1.19× &436 &\textbf{393}&1.11× & 512 &\textbf{481}  &1.06×  &  \\
480×352×97 &227 &107  &2.12×   &321  &205 &1.57×   &420 &304 &1.38×  &521 &403 &1.29×  &622 &502  &1.24×&252 &154 &1.64× &368  &\textbf{278} &1.32×  &492  &\textbf{407} &1.21×  &613 &\textbf{531} &1.15× &738 &\textbf{657} &1.12× &\\
480×352×129 &262  &135  &1.94×  &400  &263 &1.52×  & 529 &389 &1.36× &666 &518 &1.28× &794 &644 &1.23× &310  &\textbf{200} &1.55× &484  &\textbf{372} &1.30× &652 &\textbf{544} &1.20× &824 &\textbf{714} &1.15× &996 &\textbf{882} &1.13× &\\
640×352×65 &203  &92  &2.21×  &285  &170 &1.68×  &368  &250 &1.47× &491 &332 &1.48× &536 &411 &1.30×&224  &129 &1.74× &333  &\textbf{244} &1.36× &425  &\textbf{348} &1.22× &527 &\textbf{454} &1.16× &638 &\textbf{562} &1.14× &\\
640×352×97 &257  &135  &1.90×  & 388 &262 &1.48×  &522  &389 &1.34× &653&521 &1.25× &786 &643 &1.22× &303  &\textbf{206} &1.47× &473  &\textbf{373} &1.27× &650  &\textbf{545} &1.19× &820 &\textbf{713} &1.15× &983 &\textbf{883} &1.11× &\\
640×352×129 &310  &176  &1.76×  &492  &342 &1.44×  &665&\textbf{507} &1.31× &835 &\textbf{660} &1.27× &1007 &\textbf{811} &1.24× &368  &\textbf{253} &1.45× &590  &\textbf{471} &1.25× &808 &\textbf{685} &1.18× &1030 &\textbf{902} &1.14× &1240 &\textbf{1118} &1.11× &\\
800×592×65  &347  &228  &1.52×  &571  &436 &1.31×  &777  &\textbf{645} &1.20× &999 &\textbf{841} &1.19× &1205 &\textbf{1037} &1.16×&410  &\textbf{296} &1.39× &669  &\textbf{555} &1.21× &935  &\textbf{815} &1.15× &1201 &\textbf{1073} &1.12× &1471 &\textbf{1331} &1.11× &\\
800×592×97 &520  &397  &1.31×  &899  &\textbf{758} &1.19×  &1257  &\textbf{1097} &1.15× &1637 &\textbf{1433} &1.14× &1994 &\textbf{1766} &1.13×&646  &\textbf{517} &1.25× &1124 &\textbf{980} &1.15× &1609 &\textbf{1441} &1.12× &2013 &\textbf{1903} &1.06× &2570 &\textbf{2373} &1.08× &\\
800×592×129 &751  &\textbf{621}  &1.21×  &1316  &\textbf{1137} &1.16×  &1875  &\textbf{1651} &1.14× &2447 &\textbf{2174} &1.13× &3010 &\textbf{2689} &1.12×&959  &\textbf{791} &1.21× &1709 &\textbf{1515} &1.13× &2464 &\textbf{2239} &1.10× &3223 &\textbf{2964} &1.09× &3977 &\textbf{3688} &1.08× &\\
1024×576×65  &376  &251  &1.50×  &614  &\textbf{479} &1.28×  &851  &\textbf{684} &1.24× &1094 &\textbf{888} &1.23× &1323 &\textbf{1089} &1.21× &476  &\textbf{354} &1.34× &801  &\textbf{661} &1.21× &1114 &\textbf{970} &1.15× &1440 &\textbf{1280} &1.13× &1760 &\textbf{1588} &1.11× &\\
1024×576×97 &555  &430  &1.29×  &959  &\textbf{799} &1.20×  &1360  &\textbf{1144} &1.19× &1762 &\textbf{1491} &1.18× &2162 &\textbf{1832} &1.18×&731  &\textbf{591} &1.24× &1279 &\textbf{1115} &1.15× &1836 &\textbf{1639} &1.12× &2387 &\textbf{2164} &1.10× &2940 &\textbf{2689} &1.09× &\\
1024×576×129  &797  &652  &1.22×  &1402  &\textbf{1176} &1.19×  &1995  &\textbf{1698} &1.17× &2600 &\textbf{2206} &1.18× &3194 &\textbf{2726} &1.17×&1094 &\textbf{916} &1.19× &1970 &\textbf{1751} &1.13× &2848 &\textbf{2582} &1.10× &3722 &\textbf{3416} &1.09× &4599 &\textbf{4251} &1.08× &\\

  \toprule
  \multirow{3}*{Resolution}&\multicolumn{15}{c|}{\textbf{ HunyuanVideo (A6000)}}&\multicolumn{15}{c}{\textbf{HunyuanVideo (L40})}\\
    \cmidrule(lr){2-32}
  
  & \multicolumn{3}{c}{10} & \multicolumn{3}{c}{20} & \multicolumn{3}{c}{30}& \multicolumn{3}{c}{40}& \multicolumn{3}{c|}{50}&\multicolumn{3}{c}{10} & \multicolumn{3}{c}{20} & \multicolumn{3}{c}{30}& \multicolumn{3}{c}{40}& \multicolumn{3}{c}{50}\\
  \cmidrule(lr){2-4}\cmidrule(lr){5-7}\cmidrule(lr){8-10}\cmidrule(lr){11-13}\cmidrule(lr){14-16}\cmidrule(lr){17-19}\cmidrule(lr){20-22}\cmidrule(lr){23-25}\cmidrule(lr){26-28}\cmidrule(lr){29-31}
  &base&opt&spd$\uparrow$&base&opt&spd$\uparrow$&base&opt&spd$\uparrow$&base&opt&spd$\uparrow$&base&opt&spd$\uparrow$&base&opt&spd$\uparrow$&base&opt&spd$\uparrow$&base&opt&spd$\uparrow$&base&opt&spd$\uparrow$&base&opt&spd$\uparrow$
\\
  \midrule
  480×352×65 & 482 & \textbf{120} & 4.02× & 563 & \textbf{219} & 2.57× & 636 & \textbf{327} & 1.94× & 702 & \textbf{431} & 1.63× & 766 & \textbf{520} & 1.47× & 517 & \textbf{157} & 3.29× & 632 & \textbf{298} & 2.12× & 764 & \textbf{437} & 1.75× & 850 & \textbf{578} & 1.47× & 983 & \textbf{718} & 1.37×  \\
480×352×97 & 540 & \textbf{165} & 3.27× & 641 & \textbf{306} & 2.09× & 767 & \textbf{445} & 1.72× & 850 & \textbf{585} & 1.45× & 965 & \textbf{726} & 1.33×& 676 & \textbf{229} & 2.95× & 838 & \textbf{439} & 1.91× & 992 & \textbf{649} & 1.53× & 1163 & \textbf{859} & 1.35× & 1350 & \textbf{1068} & 1.26× \\
480×352×129 & 630 & \textbf{206} & 3.06× & 767 & \textbf{395} & 1.94× & 911 & \textbf{595} & 1.53× & 1071 & \textbf{791} & 1.35× & 1201 & \textbf{985} & 1.22× & 787 & \textbf{312} & 2.52× & 1038 & \textbf{602} & 1.72× & 1288 & \textbf{892} & 1.44× & 1555 & \textbf{1182} & 1.32× & 1804 & \textbf{1472} & 1.23× \\
640×352×65 & 518 & 142 & 3.65× & 596 & \textbf{269} & 2.22× & 695 & \textbf{396} & 1.76× & 791 & \textbf{515} & 1.54× & 882 & \textbf{639} & 1.38×& 601 & \textbf{195} & 3.08× & 749 & \textbf{376} & 1.99× & 902 & \textbf{555} & 1.63× & 1041 & \textbf{735} & 1.42× & 1206 & \textbf{915} & 1.32× \\
640×352×97 & 593 & \textbf{191} & 3.10× & 733 & \textbf{360} & 2.04× & 865 & \textbf{531} & 1.63× & 1008 & \textbf{701} & 1.44× & 1142 & \textbf{907} & 1.26×& 760 & \textbf{295} & 2.58× & 1021 & \textbf{571} & 1.79× & 1231 & \textbf{843} & 1.46× & 1460 & \textbf{1119} & 1.30× & 1702 & \textbf{1392} & 1.22× \\
640×352×129 & 710 & \textbf{262} & 2.71× & 899 & \textbf{503} & 1.79× & 1104 & \textbf{741} & 1.49× & 1272 & \textbf{987} & 1.29× & 1456 & \textbf{1228} & 1.19×& 987 & \textbf{409} & 2.41× & 1313 & \textbf{797} & 1.65× & 1637 & \textbf{1187} & 1.38× & 1954 & \textbf{1573} & 1.24× & 2305 & \textbf{1961} & 1.18× \\
800×592×65 & 833 & \textbf{321} & 2.60× & 1075 & \textbf{633} & 1.70× & 1294 & \textbf{942} & 1.37× & 1531 & \textbf{1252} & 1.22× & 1792 & \textbf{1561} & 1.15× & 1160 & \textbf{527} & 2.20× & 1601 & \textbf{1028} & 1.56× & 2018 & \textbf{1527} & 1.32× & 2487 & \textbf{2027} & 1.23× & 2921 & \textbf{2528} & 1.16× \\
800×592×97 & 1082 & \textbf{506} & 2.14× & 1499 & \textbf{1000} & 1.50× & 1880 & \textbf{1492} & 1.26× & 2281 & \textbf{1985} & 1.15× & 2686 & \textbf{2470} & 1.09×& 1694 & \textbf{923} & 1.84× & 2484 & \textbf{1812} & 1.37× & 3291 & \textbf{2702} & 1.22× & 4115 & \textbf{3592} & 1.15× & 4898 & \textbf{4482} & 1.09× \\
800×592×129 & 1467 & \textbf{772} & 1.90× & 2062 & \textbf{1514} & 1.36× & 2681 & \textbf{2259} & 1.19× & 3302 & \textbf{3006} & 1.10× & 3920 & \textbf{3756} & 1.04× & 2311 & \textbf{1374} & 1.68× & 3511 & \textbf{2686} & 1.31× & 4717 & \textbf{4001} & 1.18× & 5926 & \textbf{5317} & 1.11× & 7136 & \textbf{6632} & 1.08× \\
1024×576×65 & 997 & \textbf{430} & 2.32× & 1333 & \textbf{835} & 1.60× & 1676 & \textbf{1242} & 1.35× & 1989 & \textbf{1645} & 1.21× & 2324 & \textbf{2051} & 1.13× & 1486 & \textbf{760} & 1.96× & 2146 & \textbf{1490} & 1.44× & 2817 & \textbf{2223} & 1.27× & 3473 & \textbf{2952} & 1.18× & 4123 & \textbf{3682} & 1.12× \\
1024×576×97 & 1399 & \textbf{729} & 1.92× & 1987 & \textbf{1425} & 1.39× & 2545 & \textbf{2090} & 1.22× & 3150 & \textbf{2771} & 1.14× & 3726 & \textbf{3453} & 1.08×  & 2237 & \textbf{1333} & 1.68× & 3395 & \textbf{2612} & 1.30× & 4576 & \textbf{3894} & 1.18× & 5762 & \textbf{5173} & 1.11× & 6952 & \textbf{6453} & 1.08× \\
1024×576×129 & 1918 & \textbf{1097} & 1.75× & 2836 & \textbf{2078} & 1.36× & 3733 & \textbf{3090} & 1.21× & 4638 & \textbf{4206} & 1.10× & 5545 & \textbf{5240} & 1.06×& 3187 & \textbf{2024} & 1.57× & 5017 & \textbf{3984} & 1.26× & 6846 & \textbf{5942} & 1.15× & 8658 & \textbf{7899} & 1.10× & 10472 & \textbf{9856} & 1.06×  \\
  \bottomrule

\end{tabular}
}

\end{center}

\begin{center}
\makebox[\textwidth]{%
  \begin{minipage}{1.0\textwidth}
    \raggedright
    Table 6: Latency and speedup for generating 10 videos with the baseline system and our optimized PipeDiT. Bold numbers indicate the results obtained with PipeDiT w/ Aco.
  \end{minipage}
}
\end{center}

\begin{center}
\centering
\resizebox{1.0\textwidth}{!}{
\begin{tabular}{*{22}{c}}
  \toprule
  \multicolumn{21}{c}{\textbf{OpenSoraPlan(A6000)}}\\
  \cmidrule{6-21}
    \multirow{2}*{A}&\multirow{2}*{B}&\multirow{2}*{C}&\multirow{2}*{D} & \multicolumn{3}{c}{480×352×65} & \multicolumn{2}{c}{480×352×129} & \multicolumn{2}{c}{640×352×65}& \multicolumn{2}{c}{640×352×129}& \multicolumn{2}{c}{800×592×65}& \multicolumn{2}{c}{800×592×129}& \multicolumn{2}{c}{  1024×576×65}& \multicolumn{2}{c}{  1024×576×129}\\
  \cmidrule(lr){6-7}\cmidrule(lr){8-9}\cmidrule(lr){10-11}\cmidrule(lr){12-13}\cmidrule(lr){14-15}\cmidrule(lr){16-17}\cmidrule(lr){18-19}\cmidrule(lr){20-21}
  & & & & &T(s)$\downarrow$&Spd$\uparrow$&T(s)$\downarrow$&Spd$\uparrow$&T(s)$\downarrow$&Spd$\uparrow$&T(s)$\downarrow$&Spd$\uparrow$&T(s)$\downarrow$&Spd$\uparrow$&T(s)$\downarrow$&Spd$\uparrow$&T(s)$\downarrow$&Spd$\uparrow$&T(s)$\downarrow$&Spd$\uparrow$ \\
  \midrule
  \rowcolor{gray!15}\noalign{\vskip -3.pt}
\ding{51}&\ding{55}&\ding{55}&\ding{55}& &314&1×&529&1×&368&1×&665&1×&777&1×&1875&1×&851&1×&1995&1×\\

\ding{55}&\ding{51}&\ding{55}&\ding{55}&&217&1.45×&452&1.17×&234&1.57×&500&1.33×&649&1.20×&1872&1.00×&702&1.21×&2138&0.93×\\
\ding{55}&\ding{51}&\ding{51}&\ding{55}&&200&1.57×&390&1.36×&250&1.47×&509&1.31×&649&1.20×&1847&1.02×&717&1.19×&1936&1.03×\\
\ding{55}&\ding{51}&\ding{51}&\ding{51}&&261&1.20×&414&1.28×&296&1.24×&507&1.31×&645&1.20×&1652&1.14×&683&1.25×&1690&1.18×\\

  \toprule
  \multicolumn{21}{c}{\textbf{OpenSoraPlan(L40)}}\\
  \midrule
  \rowcolor{gray!15}\noalign{\vskip -3.pt}
  \ding{51}&\ding{55}&\ding{55}&\ding{55}& &355&1×&652&1×&425&1×&808&1×&935&1×&2464&1×&1114&1×&2848&1×\\
\ding{55}&\ding{51}&\ding{55}&\ding{55}&&274&1.30×&609&1.07×&372&1.14×&792&1.02×&962&0.97×&2816&0.88×&1161&0.96×&3257&0.87×\\
\ding{55}&\ding{51}&\ding{51}&\ding{55}&&299&1.19×&621&1.05×&372&1.14×&789&1.02×&963&0.97×&2819&0.87×&1158&0.96×&3231&0.88×\\
\ding{55}&\ding{51}&\ding{51}&\ding{51}&&298&1.19×&544&1.20×&348&1.22×&685&1.18×&815&1.15×&2239&1.10×&970&1.15×&2582&1.10×\\
\toprule
\multicolumn{21}{c}{\textbf{HunyuanVideo(A6000)}}\\
  \midrule
  \rowcolor{gray!15}\noalign{\vskip -3.pt}
  \ding{51}&\ding{55}&\ding{55}&\ding{55}& &636&1×&911&1×&695&1×&1104&1×&1294&1×&2681&1×&1676&1×&3733&1×\\
\ding{55}&\ding{51}&\ding{55}&\ding{55}&&340&1.87×&681&1.34×&403&1.72×&824&1.34×&984&1.32×&2501&1.07×&1374&1.22×&3680&1.01×\\
\ding{55}&\ding{51}&\ding{51}&\ding{55}&&345&1.84×&701&1.30×&404&1.72×&824&1.34×&983&1.32×&2499&1.07×&1374&1.22×&3675&1.02×\\
\ding{55}&\ding{51}&\ding{51}&\ding{51}&&327&1.94×&595&1.53×&396&1.76×&741&1.49×&942&1.37×&2259&1.19×&1242&1.35×&3090&1.21×\\
\toprule
\multicolumn{21}{c}{\textbf{HunyuanVideo(L40)}}\\
  \midrule
  \rowcolor{gray!15}\noalign{\vskip -3.pt}
  \ding{51}&\ding{55}&\ding{55}&\ding{55}& &764&1×&1288&1×&902&1×&1637&1×&2018&1×&4717&1×&2817&1×&6846&1×\\
\ding{55}&\ding{51}&\ding{55}&\ding{55}&&466&1.64×&1087&1.18×&600&1.50×&1380&1.19×&1690&1.19×&4751&0.99×&2443&1.15×&7180&0.95×\\
\ding{55}&\ding{51}&\ding{51}&\ding{55}&&468&1.63×&1086&1.19×&599&1.51×&1380&1.19×&1687&1.20×&4749&0.99×&2441&1.15×&7175&0.95×\\
\ding{55}&\ding{51}&\ding{51}&\ding{51}&&437&1.75×&892&1.44×&555&1.63×&1187&1.38×&1527&1.32×&4001&1.18×&2223&1.27×&5942&1.15×\\
\bottomrule

\end{tabular}
}

\end{center}

\noindent
\makebox[\textwidth][c]{%
    \parbox{1\textwidth}{
        \centering
        Table 7: Efficiency improvement of different optimization methods.
    }
}
\subsection{Consistency Proof of Generated Results}
Since our optimization focuses solely on resource allocation and computational workload balancing, the algorithmic logic remains fully consistent with the original method. As a result, the generated outputs are identical to those of the original algorithm. Fig. 6 presents the outputs from both the original method and our optimized approach under the same prompt, experimental configuration, and sampled frame index. The results clearly demonstrate that the two generations are perfectly identical.

\clearpage

\begin{figure}[!t]
\centering
\includegraphics[width=1.0\textwidth]{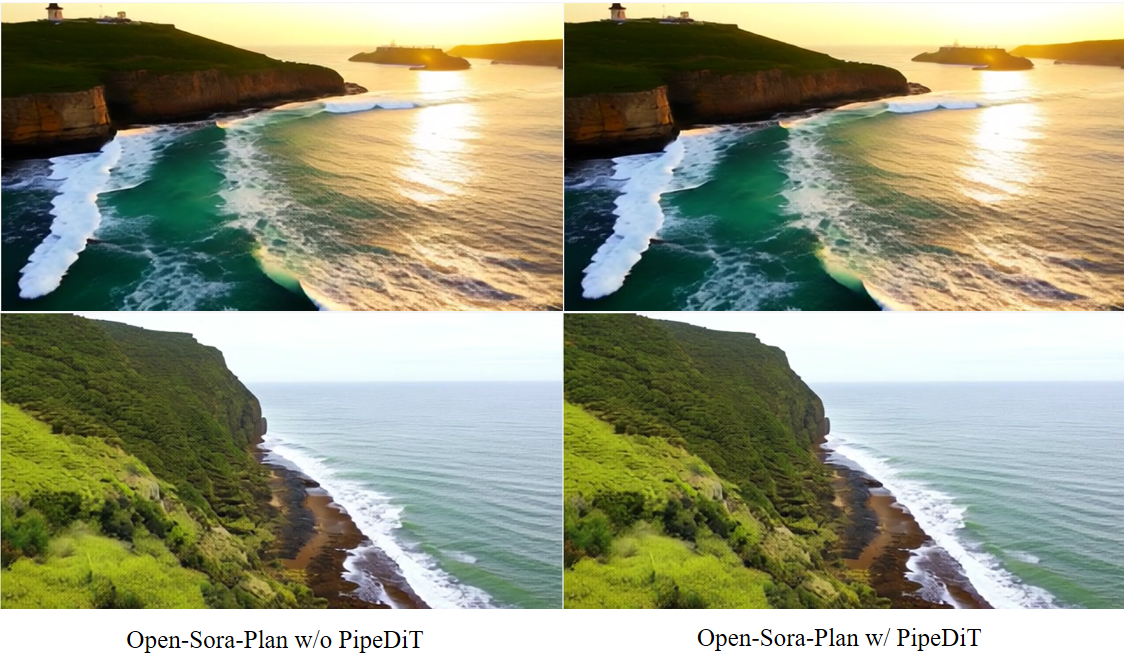}
\label{fig:aco2}
\end{figure}

\makebox[\textwidth][c]{%
    \parbox{1\textwidth}{
        \centering
        Figure 6: The generation results show that the outputs produced by PipeDiT are consistent with those of the original algorithm.
    }
}

\end{document}